\documentclass[letterpaper]{article} % DO NOT CHANGE THIS
\usepackage{aaai25}  % DO NOT CHANGE THIS
\usepackage{times}  % DO NOT CHANGE THIS
\usepackage{helvet}  % DO NOT CHANGE THIS
\usepackage{courier}  % DO NOT CHANGE THIS
\usepackage[hyphens]{url}  % DO NOT CHANGE THIS
\usepackage{graphicx} % DO NOT CHANGE THIS
\usepackage{natbib}  % DO NOT CHANGE THIS AND DO NOT ADD ANY OPTIONS TO IT
\usepackage{caption} % DO NOT CHANGE THIS AND DO NOT ADD ANY OPTIONS TO IT
\usepackage{algorithm}
\usepackage{algorithmic}

\usepackage{newfloat}
\usepackage{listings}
\DeclareCaptionStyle{ruled}{labelfont=normalfont,labelsep=colon,strut=off} % DO NOT CHANGE THIS
\floatstyle{ruled}
\newfloat{listing}{tb}{lst}{}
\floatname{listing}{Listing}
\usepackage{xcolor}
\newcommand{\answerYes}[1]{\textcolor{blue}{#1}} 
\newcommand{\answerNo}[1]{\textcolor{teal}{#1}} 
\newcommand{\answerNA}[1]{\textcolor{gray}{#1}}

\usepackage{amsmath}

\usepackage[np]{numprint}
\usepackage{booktabs}
\usepackage{siunitx}

\usepackage{xcolor}

\usepackage{xspace}

\newcommand{\fsl}{\textsl}

\usepackage[most]{tcolorbox}
\usepackage{multirow}

\usepackage{mathtools}
\usepackage{booktabs}
\usepackage{url}
\usepackage{tabularx}
\usepackage{graphicx}
\usepackage{subcaption}
\usepackage{pifont}
\usepackage{tikz}
\usepackage{standalone}
\usepackage{pgfplots}
\pgfplotsset{compat=1.18}
\usepgfplotslibrary{groupplots}
\usepackage{pgfplotstable}
\usepackage{tikz-dependency}
\usepackage[T1]{fontenc}
\usetikzlibrary{fit}
\usetikzlibrary{positioning}

\pgfdeclarelayer{bg}    % declare background layer
\pgfsetlayers{bg,main}  % set the order of the layers (main is the standard layer)
\definecolor{oceanboatblue}{rgb}{0.0, 0.47, 0.75}
\definecolor{goldenpoppy}{rgb}{0.99, 0.76, 0.0}
\definecolor{brickred}{rgb}{0.8, 0.25, 0.33}
\definecolor{asparagus}{rgb}{0.53, 0.66, 0.42}
\definecolor{mangotango}{rgb}{1.0, 0.51, 0.26}

\title{A Benchmark for Cross-Domain Argumentative Stance Classification\\ on Social Media}

\author{
    Jiaqing Yuan,
    Ruijie Xi,
    Munindar P. Singh
    \\    
}
\affiliations{   
    North Carolina State University \\
    Raleigh, North Carolina 27606 USA\\
    jyuan23@ncsu.edu, rxi@ncsu.edu, mpsingh@ncsu.edu
}

\begin{document}

\maketitle
\thispagestyle{plain}
\pagestyle{plain}

\begin{abstract}
Argumentative stance classification plays a key role in identifying authors' viewpoints on specific topics. However, generating diverse pairs of argumentative sentences across various domains is challenging. Existing benchmarks often come from a single domain or focus on a limited set of topics. Additionally, manual annotation for accurate labeling is time-consuming and labor-intensive. To address these challenges, we propose leveraging platform rules, readily available expert-curated content, and large language models to bypass the need for human annotation. Our approach produces a multidomain benchmark comprising 4,498 topical claims and 30,961 arguments from three sources, spanning 21 domains. We benchmark the dataset in fully supervised, zero-shot, and few-shot settings, shedding light on the strengths and limitations of different methodologies.
We release the dataset and code in this study at \textit{hidden for anonymity}. 

\end{abstract}

\section{Introduction}
Argumentation is a pervasive human activity present in various aspects of everyday life, which involves expressing viewpoints backed by reasons or attempting to persuade others towards a specific perspective \citep{sobhani-etal-2015-argumentation}. In the context of argument mining, a crucial task is argumentative stance classification \citep{10.1145/3369026}, where the goal is to classify an argument's stance as either \emph{favor}, \emph{against}, or \emph{neutral} regarding a given claim. 
For example, argument \fsl{The possession of nuclear weapons provides countries with a strong defense mechanism, deterring potential adversaries from launching attacks} can be classified as \emph{against} the claim \fsl{All countries should give up their nuclear weapons}.

In recent years, social media platforms have become effective channels for sharing information, encouraging diverse perspectives, and facilitating the exchange of ideas \citep{ALDAYEL2021102597}. Claims are often simplified into noun phrase topics, such as ``nuclear weapon'' for the above example. Previous research has spent a lot of effort in constructing datasets concerning various topics. For example, \citet{mohammad-etal-2016-semeval} constructed a dataset with tweets commenting on \fsl{Atheism}, \fsl{Climate change}, \fsl{Feminist}, \fsl{Hillary Clinton}, \fsl{Abortion}, and \fsl{Donald Trump}. \citet{conforti-etal-2020-will} studied public opinion toward four financial merger events, \citet{glandt-etal-2021-stance} investigated stance classification on three policies during Covid-19, and most recently, \citet{cruickshank2024diversedatasetyoutubevideo}  collected YouTube video comments and annotated for their stance towards the U.S. military. 

One challenge to stance classification comes from the variety of stance topics \citep{allaway-mckeown-2020-zero}. Many prior benchmarks are not topically diverse. As mentioned above, they typically feature a handful of topics, each with a corpus of comments to facilitate the training of supervised models \emph{dedicated} to that topic \citep{stab-etal-2018-cross}. The acquisitions of stance labels rely on human annotators for ground truth \citep{10.1145/3369026}, which is time-consuming and difficult to scale up. Besides, most previous benchmarks focus on a single genre or source. 

\begin{table*}[t]
\centering
\scalebox{1}{%
\begin{tabular}{p{3cm} p{12.5cm} l }
\toprule
Topical Claim & Comment & Stance \\ 
 \midrule
Space exploration is a waste of money.& Instead of decreasing resources by space travel and such, we must deal with problems on Earth first. Why bother spending all this money on exploring space when we could be helping our own planet that us humans live on\ldots & Favor\\
Animals have rights.&It makes no sense to give animals rights because they cannot makes decisions about what is right and wrong and will not try to treat us in an ethical manner in return\ldots & Against\\
All student loan debt should be eliminated.&Schools already have a heavy workload and limited resources. Adding moral education to their curriculum may place an additional burden on teachers and administrators. It could divert valuable time and resources away from core academic subjects, potentially compromising the quality of education provided to students\ldots & None\\
 %\midrule
 \bottomrule
\end{tabular}
}
\caption{Examples from our benchmark.}
\label{table:examples}
\end{table*}

Accordingly, our objective is to construct a diverse and multisource stance classification benchmark without human annotation. 
\citet{allaway-mckeown-2020-zero} categorize stance classification into two categories based on the topic: \emph{topic-phrase} and \emph{topic-position}. 
For the former, the topic is typically a noun phrase (including proper noun), such as \fsl{nuclear weapon}. 
For the latter, the topic is a complete position claim such as \fsl{All countries should give up their nuclear weapons}. 
Notably, the argument we introduce at the beginning of this section would be classified as \emph{favor} for the former and \emph{against} for the latter. This reveals a significant difference between topic-phrase and topic-position stance classification: the latter is context-dependent. Our benchmark focuses on topic-position, as we argue that topic-phrase can be easily converted into topic-position by constructing a positional claim, which has a more general form. Preferably, a truly intelligent stance classification system should be able to grasp the meaning of the topical claim and reverse its prediction when the claim reverses itself.

We construct our benchmark from three distinctive sources: a social media platform, two debate websites, and arguments generated by large language models (LLMs). For the social media platform, we leverage conversations from a subreddit called ChangeMyView\footnote{https://www.reddit.com/r/changemyview/} from Reddit, where a poster challenges other users to change the poster's opinion expressed by a positional title. The comments labeled by the poster as successful can be seen as counterarguments to the title. On the debate websites, opposing arguments are curated by the authors, who provide clear stance labels. LLMs have been shown effective for data augmentation \citep{sahu-etal-2022-data, yoo-etal-2021-gpt3mix-leveraging, edwards-etal-2022-guiding}. To further enrich the diversity of the benchmark, we leverage LLMs to generate arguments on both sides for a given topical claim. The foregoing steps yield a combined dataset of \np{4,498} topics and \np{30,961} arguments spanning 21 domains. Table~\ref{table:comparison} compares our benchmark with previous benchmarks. 

With this benchmark, we aim to answer the following research questions. 

\begin{description}

    \item[RQ1] \textbf{How does LLMs generated data benefit stance classification in real-world applications?} 
    
    We conduct experiments on individual real-world datasets and on combined LLM-generated and real-world datasets to show the benefit of integrating LLM-generated data during training. We use traditional machine learning methods \citep{aldayel-2021-stance} commonly applied in stance classification alongside pretrained LLMs like BERT \cite{devlin-etal-2019-bert}. 
    
    \item[RQ2] \textbf{To what extent do stance classification models generalize across various topics and domains within a topic-position framework?} 
    
    As with the above discussion, topic-position stance classification offers a notable advantage due to its flexibility. It analyzes pairs of argumentative sentences instead of being limited to the topical noun phrase, as seen in topic-phrase setting. This research question aims to explore how effectively stance classification can generalize across both different topics—specifically, those sourced from the same platform yet covering distinct topics—and across, those sourced from different platforms.
    
    \item[RQ3] \textbf{How does supervised finetuning compare to zero-shot and few-shot learning with LLMs for cross-domain stance classification?} 
    
    Supervised finetuning and in-context learning are two common methods for adapting models to specific tasks. Research has shown that LLMs can adapt well to new tasks in zero-shot and few-shot scenarios \cite{NEURIPS2020_1457c0d6}, which do not require any training data. In contrast, supervised finetuning significantly improves performance on data within the trained domain, but it often fails to generalize effectively to new domains \cite{NG2022103070}. This study aims to examine the relative effectiveness of these two approaches in the context of cross-domain stance classification.

\end{description}

\paragraph{Findings}
In our analysis of traditional machine learning techniques, including Support Vector Machines (SVM), Convolutional Neural Networks (CNN), and Bidirectional Long Short-Term Memory (BiLSTM) networks, we observe that incorporating LLMs generated data into the training process significantly enhances in-domain performance for these models. However, this strategy yields inconsistent outcomes when applied to finetuning contemporary LLMs. We also observe that generative models consistently outperform classification models with supervised finetuning. LLMs display commendable performance in zero-shot settings for cross-domain evaluation, though a substantial performance gap remains in comparison to in-domain supervised finetuning. Furthermore, in few-shot experiments, instruction-tuned LLMs consistently outperform their non-instruction-tuned counterparts, highlighting the effectiveness of instruction-tuning as a robust approach for adapting LLMs to downstream tasks.

\paragraph{Contributions}
Our contributions are twofolds:

\begin{itemize}
\item We propose a scalable and extensible framework to construct a diverse and multisource benchmark for argumentative stance classification without involving human annotation. 

\item We implement and evaluate fully-supervised learning, zero-shot learning, and few-shot learning using LLMs. This thorough assessment facilitates a comparative analysis of various methodologies, emphasizing the efficacy of instruction-tuning for optimizing the performance of LLMs.
\end{itemize}

\begin{table*}[t]
\centering
\scalebox{1}{%
\begin{tabular}{l l l l r}
\toprule
Authors & Diversity & Source & Topic Type & Size \\ 
\midrule
\citet{mohammad-etal-2016-semeval} & 6 topics & Twitter &Phrase & 4,870\\
\citet{stab-etal-2018-cross}&8 topics & Google query & Phrase & 25,492\\
\citet{allaway-mckeown-2020-zero} & 4,641 noun phrases & Debate website & Phrase & 18,545\\
\citet{ferreira-vlachos-2016-emergent} & 300 rumor claims & News article & Position & 2,595\\
\citet{gorrell-etal-2019-semeval} & various claims & Twitter, Reddit & Position & 8,574\\
\citet{bar-haim-etal-2017-stance} & 55 claims & Debate website & Position & 2,394\\
\citet{hanselowski-etal-2019-richly} & 6,422 claims & Fact-check website & Position & 19,439\\
\textbf{Ours} & 4,498 claims & Reddit, Debate website, LLM & Position & 30,961\\
\bottomrule
\end{tabular}
}
\caption{Comparison with previous benchmarks.}
\label{table:comparison}
\end{table*}    

The rest of this paper is structured as follows: Section~\ref{sec: related work} reviews related work and positions our research in relation to existing studies. Section~\ref{sec: benchmark construction} introduces the proposed framework for constructing a multisource stance classification benchmark. In Section~\ref{sec:supervised finetuning}, we present experiments using both traditional machine learning models and contemporary LLMs with supervised finetuning. Section~\ref{sec:llm evaluation} examines zero-shot and few-shot learning across various LLM families. Section~\ref{sec:conclusion} provides a summary of our findings and presents the concluding remarks. Lastly, Section~\ref{sec:discussion} discusses the broader impact of this work.

\section{Related Work}
\label{sec: related work}

Stance is a speaker's evaluation of a proposition or topic. 
The proposition may be implicit as a \emph{topic-phrase} (a noun phrase) or explicit as a \emph{topic-position} (a positional claim) \citep{allaway-mckeown-2020-zero}. 
Datasets from early research originate from arguments in online debate forums \citep{somasundaran-wiebe-2010-recognizing, murakami-raymond-2010-support, walker-etal-2012-corpus, hasan-ng-2014-taking} and mostly fall under the topic-phrase category \citep{ALDAYEL2021102597}. 
More recent datasets cover various topics \citep{sobhani-etal-2017-dataset, qazvinian-etal-2011-rumor, mohammad-etal-2016-semeval, conforti-etal-2020-will, li-etal-2021-p, glandt-etal-2021-stance}. For topic-position stance classification, datasets primarily come from news articles, where headlines are used as the topic-phrase \citep{ferreira-vlachos-2016-emergent, habernal-etal-2018-argument, conforti-etal-2020-stander, chen-etal-2019-seeing, qazvinian-etal-2011-rumor}. 
Many existing datasets are generated from one source in one domain and focus on comments for a small set of topics, followed by human annotation. 

We emphasize the topic-position variant of stance classification because phrases can be transformed into positions by formulating an affirmative claim (e.g., \fsl{Abortion} maps to \fsl{Abortion should be legalized}). 
Unlike previous works, which rely on human annotators for labeling, we leverage platform rules, readily available expert-curated content, and large language models to acquire faithful stance labels. 

\section{Benchmark Construction}
\label{sec: benchmark construction}

We now describe the details for building our benchmark, which includes \np{4,498} topical claims, and \np{30,961} arguments, covering 21 domains and three distinct sources. 

\subsection{Dataset Collection}
In order to enhance textual diversity, our benchmark is curated with content from three types of sources: a social media platform, debate websites, and arguments generated by a large language model.

\paragraph{Social media platform}

ChangeMyView (CMV) is a subreddit from Reddit that serves as a dedicated forum for fostering discourse. Within this subreddit, participants actively contribute their opinions and engage in discussions with the explicit aim of defending their perspectives. A typical CMV post adheres to a particular structure: it begins with the abbreviation ``CMV: '' signifying Change My View, followed by a concise representation of the author's viewpoint. Subsequently, the body of the post features a comprehensive elucidation by the author, providing an in-depth account of the reasoning and rationale supporting their stance. If any of the comments made by the participants successfully manage to influence a shift in the author's viewpoint, the author acknowledges this change by awarding the commenter with a \emph{delta}. Therefore, we extract the title, body, and delta-awarded comments, where the title corresponds to the designated topical claim, the body constitutes a supportive argument, and the comments bearing a delta reward act as counterarguments. 

We leverage two existing CMV datasets \citep{10.1145/2872427.2883081, alkhatib:2020b} due to Reddit's recent limits on the API (at most the most recent 1000 posts are returned by an API call). To keep the text meaningful and concise, we select bodies and comments with a length between 20 to 200 words. 

\paragraph{Debate websites} 
We selected two online platforms dedicated to debates between users, \url{idebate.net} and \url{debatewise.org}, due to their well-structured presentation of arguments for and against a position. These platforms provide clear and comprehensive arguments, thus obviating the need for annotation. We captured the subjects of a debate along with the associated arguments.

Some of the arguments were excessively verbose. 
However, we observed that the stance of an argument can typically be discerned within the initial few sentences. Therefore, in the interest of improving the manageability of the data, we retained only the initial five sentences of each argument. This choice aligns with our restricting the CMV arguments to 20 to 200 words.

\begin{figure}[ht]
\centering
\includegraphics[width=1\columnwidth]{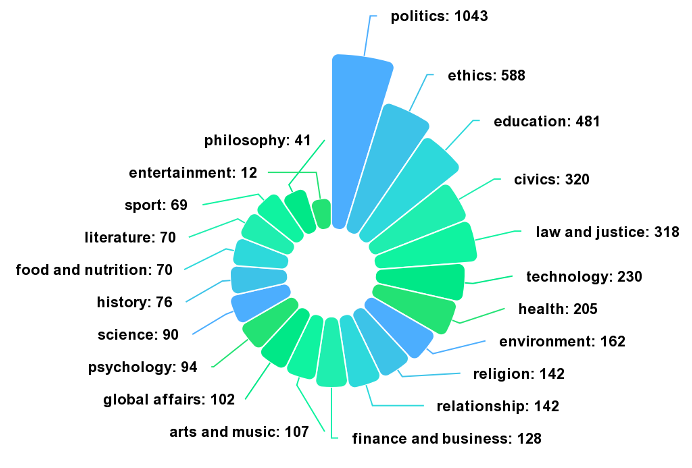}
\caption{Domain distribution of the topics in our proposed dataset.}
\label{fig:domains}
\end{figure}

\subsection{Large Language Model Generated Argument}
\paragraph{Generating arguments using GPT-3}

We use GPT-3 to generate text because it produces coherent, contextually relevant, and fluent language. 
GPT models are pretrained on vast amounts of diverse data, enabling them to understand and mimic human-like language patterns across various topics. 
Previous research has demonstrated GPT's ability to encode beliefs into argumentative texts \citep{alshomary-2021-belief} and to determine rumors and antagonistic relationships between Twitter users by detecting stances in replies and quotes \citep{villa-2020-stance}. 
In the context of this study, we employ GPT-3\footnote{https://platform.openai.com/docs/guides/text-generation/chat-completions-api} to produce arguments from multiple perspectives on diverse subjects. This process involves three steps.

\paragraph{Topic collection}
We conducted a search with Google to identify contentious subjects and compiled such topics from ten web sources (listed in Appendix~\ref{app:source}).

\paragraph{Claim generation}
The topics collected from the web have various forms. We use Prompt~1 shown below to generate a claim for each topic. 

\begin{tcolorbox}[colback=black!5!white,colframe=gray!150!black,title=Prompt 1, coltitle = black!20!black]
  Given a question or topic, generate a controversial claim.

Input: Should Halloween costumes be allowed in schools?

Output:
\end{tcolorbox}

\paragraph{Argument generation}
For each claim, we use Prompt~2, shown below, to generate three arguments from both sides. 

\begin{tcolorbox}[colback=black!5!white,colframe=gray!150!black,title=Prompt 2, coltitle = black!20!black]
Given a topic, write three distinct supporting arguments and three opposing arguments. You should write in 1st person view rather than 3rd person view.  Don't explicitly say I support or oppose. Don't summarize the points at the beginning.

Topic: Halloween costumes should be allowed in schools.
\end{tcolorbox}

\begin{table}[!htb]
\centering
\scalebox{1}{%
\begin{tabular}{l r r r r}
\toprule
 & CMV & Debate & LLMG& Total\\ 
 \midrule
 Topical claims & 1,873 & 982 & 1,643 & 4,498 \\
 Arguments & 6,407 & 14,696 & 9,858 & 30,961 \\
 Favor & 1,863 & 7,319 & 4,929 & 14,111\\
 Against & 4,544 & 7,377 & 4,929 & 16,850\\
 None & 5,609 & 2,904 & 4,929 & 13,442\\
 Arg/Claim & 6.4 & 17.9 & 9.0 & 6.9 \\
 Length/Arg & 105.2 & 102.3 & 56.2 & 86.7\\
 %\midrule
 \bottomrule
\end{tabular}
}%
\caption{Statistics for our proposed benchmark. \#Arg/Claim means the average number of arguments per claim, and \#Length/Arg means the average number of words for each argument.}
\label{table:statistics}
\end{table}

\begin{tcolorbox}[colback=black!5!white,colframe=gray!150!black,title=Prompt 3, coltitle = black!20!black]
Given a topic, classify which domain the topic falls into. Output the domain directly without other words. Some example domains are sport, environment, civics, history, education, politics, technology, literature, arts and music, science, ethics and animal, finance and business, global affairs, health, psychology, law and justice, relationship, nursing, religion, food and nutrition. You should pick the category that most closely matches the topic. If none of the categories matches, you can use a new category of your own.

Topic: Halloween costumes should be allowed in schools.

Output:
\end{tcolorbox}

\paragraph{Constructing neutral arguments}
The above methods assign stance labels of favor or against, as in some prior datasets. 
Generating a neutral stance, however, is difficult since a judgment of neutrality often depends on each annotator's interpretation. 
We seek to compel the LLM to focus on how the argument and the topical claim relate, moving beyond reliance on surface-level linguistic cues. 

For each claim, We create neutral arguments by randomly selecting arguments for other claims. Consequently, we define a neutral stance as one that includes either irrelevant arguments or instances where no discernible stance can be inferred. One way is to randomly sample arguments. However, it may yield instances with distinct semantics that are easily captured by the model, that is, the sampled arguments address completely different topic from the claim, which leads to easy samples. To improve this, we use BERT to embed all claims and arguments. For each claim, we randomly sample three arguments falling within the similarity score range of $[0.3,0.5]$. 
This criterion is motivated by the fact that highly similar arguments may include content that may convey an implied stance. 
Conversely, moderately similar arguments may seem to discuss related subjects but be subtly different, thereby forming more challenging examples. For example, the third claim in Table~\ref{table:examples} concerns student loan debt, but the comment is about moral education: thus it doesn't indicate a stance about the claim though they are both education.

\begin{table*}[!htb]
\centering
\scalebox{0.72}{%
\begin{tabular}{l|rrrrrrrrr|rrrrrrrrr}
\toprule
\multirow{2}{*}{\textbf{Model}} & \multicolumn{3}{c}{\textbf{CMV}} & \multicolumn{3}{c}{\textbf{Debate}} & \multicolumn{3}{c|}{\textbf{LLMG}}& \multicolumn{3}{c}{\textbf{CMV + LLMG}} & \multicolumn{3}{c}{\textbf{Debate + LLMG}} & \\
\cmidrule(lr){2-4} \cmidrule(lr){5-7} \cmidrule(lr){8-10} \cmidrule(lr){11-13} \cmidrule(lr){14-16} \cmidrule(lr){17-19}
 & {Against} & {Favor} & {None} & {Against} & {Favor} & {None} & {Against} & {Favor} & {None} & {Against} & {Favor} & {None} & {Against} & {Favor} & {None}\\
\midrule
SVM & $0.423$ & $0.142$ & $0.332$ & $0.311$ & $0.340$ & $0.168$ & $0.520$ & $0.568$& $0.558$ & $0.463$~\textsubscript{${4.0}$} & $0.332$~\textsubscript{${19.0}$} & $0.426$~\textsubscript{${16.9}$} & $0.480$~\textsubscript{${9.4}$} & $0.492$~\textsubscript{${15.2}$} & $0.265$~\textsubscript{${9.7}$}\\
CNN & $0.481$ & $0.328$ & $0.473$ & $0.421$ & $0.460$ & $0.237$ & $0.620$ & $0.646$ & $0.598$ & $0.546$~\textsubscript{${6.5}$} & $0.434$~\textsubscript{${15.0}$} & $0.554$~\textsubscript{${8.1}$} & $0.552$~\textsubscript{${13.1}$} & $0.506$~\textsubscript{${5.6}$} & $0.286$~\textsubscript{${4.9}$}\\
BiLSTM & $0.524$ & $0.372$ & $0.531$ & $0.471$ & $0.474$ & $0.356$& $0.662$ & $0.645$ & $0.651$ & $0.564$~\textsubscript{${4.0}$} & $0.450$~\textsubscript{${16.9}$} & $0.585$~\textsubscript{${7.8}$} & $0.526$~\textsubscript{${5.4}$} & $0.510$~\textsubscript{${3.6}$} & $0.386$~\textsubscript{${3.0}$}\\
\bottomrule
\end{tabular}
}%
\caption{Macro-F1 scores with a single dataset versus using LLMG for weak supervision. The subscript numbers indicate performance improvement compared to a single dataset. The training and test splits are consistent with those in Table~\ref{table:splits}. The test set for CMV+LLMG and Debate+LLMG are the same as merely training using CMV and Debate, where the training set merges CMV and Debate with LLMG.}
\label{table:baseline}
\end{table*}

\begin{table}[!htb]
\centering
\begin{tabular}{l l r r r}
\toprule
\textbf{Model} & \textbf{Dataset} & \textbf{Against} & \textbf{Favor} & \textbf{None} \\ 
\midrule
\multirow{2}{*}{\textbf{SVM}} & CMV    & 0.40**  & 1.90*** & 0.94**  \\  
& Debate & 0.40**  & 1.20**  & 0.67**  \\  
\midrule
\multirow{2}{*}{\textbf{CNN}} & CMV    & 0.65**  & 1.06**  & 0.81*** \\  
& Debate & 0.75*** & 1.31**  & 0.49**  \\  
\midrule
\multirow{2}{*}{\textbf{BiLSTM}} & CMV    & 0.40**  & 0.78*** & 0.54**  \\  
   & Debate & 0.68**  & 0.91*** & 0.42**  \\  
\bottomrule
\end{tabular}
\caption{Cohen's \( d \) Effect Sizes for SVM, CNN, and BiLSTM Models on the three datasets with and without LLMG. ** and *** denote varying levels of statistical significance, where ** means a $p\-value < 0.05$ and ** means a $p\-value < 0.001$. The p-values are computed using a t-test to compare the means of Macro-F1 scores between the two groups.}
\label{tab:cohens_d}
\end{table}

\subsection{Dataset Characteristics}
We use ``LLMG'' to refer to the dataset generated by GPT-3. Table~\ref{table:examples} and Table~\ref{table:statistics} show some examples and the statistics for our datasetively. Our dataset exhibits greater diversity than prior datasets. We apply GPT-3 to classify them into predefined categories as shown by Prompt~3. Furthermore, we allow GPT-3 to generate novel domains. This process yields over 100 domains across all topics. 
We consolidate these into 21 principal domains. 
Figure~\ref{fig:domains} illustrates the distribution of these domains across the topics.

\subsection{Evaluate LLM Generated Dataset}

To evaluate our LLM-generated dataset's (LLMG) quality, we (1) manually verify that GPT-3 adheres to the guidelines and produces accurate responses and (2) compare the lexical diversity of real-world datasets and GPT-3 generated content.

\paragraph{Human verification}
First, we applied regular expressions to search for phrases like \fsl{as an AI, I cannot} and its variations, such as \fsl{as an AI, I can't}, in LLMG and found no such occurrences. 
Second, three independent raters labeled 200 randomly selected arguments from LLMG.

We designed a survey to assess two key aspects: (1) AI acceptance---whether GPT-3 refused to respond, and (2) AI accuracy---whether the response contained three supporting and three opposing arguments in the desired order. This survey included two evaluation questions:

\begin{description}
    \item[AI acceptance:] Does the sentence indicate that the AI refused to provide a response? For instance, does it contain variations of \fsl{as an AI, I cannot}?
    \item[AI accuracy:] Does the response adhere to Prompt~2, specifically including three supporting and three opposing arguments in the correct order?
\end{description}

Each question could be answered as \emph{Yes} or \emph{No}. 
We observed that the raters, based on majority voting, found no instances of refusal to answer, and 99\% of the responses adhered to the instructions in Prompt~2. 
The few answers that did not follow the prompt were instances where the first-person perspective was not used. 
Both tasks were rated as \emph{strongly related}, with Cohen's Kappa scores of $1.0$ for AI acceptance and $0.95$ for AI accuracy. These results indicate that LLMG contains high-quality text.

\begin{figure}[!htb]
    \centering
    \begin{subfigure}{\columnwidth}
\includegraphics[scale=0.46]{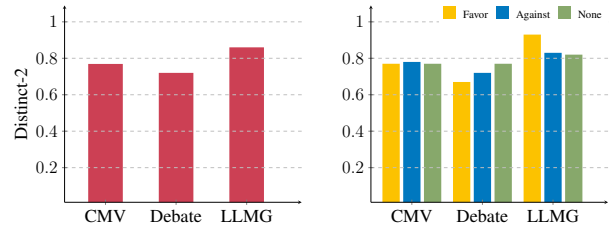}
\caption{Lexical diversity is calculated by bigram percentage.}
\label{fig:voc_diversity}
    \end{subfigure}
    \begin{subfigure}{\columnwidth}
\includegraphics[scale=0.46]{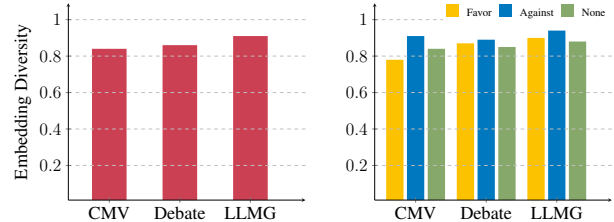}
\caption{Semantic diversity is calculated as $(1-\sigma)$, where $\sigma$ is the relevant cosine similarity between EBR embeddings.}
\label{fig:sem_diversity}
    \end{subfigure}
    \caption{Lexical and semantic diversity scores across datasets and labels.}
\end{figure}

\paragraph{Lexical and semantic diversity}
Figure~\ref{fig:voc_diversity} illustrates that sentences in LLMG exhibit greater lexical diversity than those in the CMV and Debate datasets. 
Lexical diversity is quantified using the metric called \emph{distinct-2}---the number of unique bigrams and normalizing by the total number of words generated---which is a popular metric for lexical diversity \citep{li-2016-diversity, park-2019-generating}.

The overall distinct-2 scores---i.e., averaged across all sentences on the three labels---are as follows: $0.774$ for CMV, $0.719$ for Debate, and $0.863$ for LLMG. Specifically, the distinct-2 scores for Favor are $0.771$ for CMV, $0.782$ for Debate, and $0.767$ for LLMG; for Against; they are $0.674$ for CMV, $0.719$ for Debate, and $0.770$ for LLMG; and for None, they are $0.792$ for CMV, $0.825$ for Debate, and $0.823$ for LLMG. 
This finding aligns with previous research indicating that machine-generated text often exhibits greater lexical diversity than human-authored text \citep{lee-2022-coauthor, ravi-2024-small}.

Moreover, we compare semantic diversity across the three datasets. Similarity is determined by averaging the cosine similarity between the BERT embeddings of each instance. Semantic diversity is measured by averaging the cosine similarities between BERT embeddings for each instance. A lower similarity indicates higher diversity.
Figure~\ref{fig:sem_diversity} illustrates that the overall semantic diversity values are $0.84$ for CMV, $0.86$ for Debate, and $0.91$ for LLMG. This indicates that LLMG has the highest semantic diversity of the three datasets. 
Breaking down by label for CMV, Debate, and LLMG, respectively, we see that: for Favor, the diversity values are $0.78$, $0.87$, and $0.90$; for Against, $0.91$, $0.89$, and $0.94$; and for None, $0.84$, $0.85$, and $0.88$.

In other words, LLMG shows the highest diversity across all labels, underscoring its superiority in capturing greater lexical and semantic variations than both CMV and Debate. Thus, LLMG is potentially better suited for more nuanced analyses than the other datasets, making it a valuable resource for investigating stance.

\begin{table*}[t]
\centering
\scalebox{1}{
\begin{tabular}{l r r r r r r r r r r r r }
\toprule
 & \multicolumn{4}{c}{train} & \multicolumn{4}{c}{dev} & \multicolumn{4}{c}{test}\\ 
 \cmidrule(lr){2-5} 
 \cmidrule(lr){6-9} 
 \cmidrule(lr){10-13} 
 & topics & args &favor & against& topics & args &favor&against& topics & args&favor & against\\ 
 \midrule
 CMV & 748 & 2,633 &744&1,889& 188 & 626 &188&438& 937 & 3,148&931&2,217\\ 
 Debate & 392 & 5,860 &2,934&2,926& 99 & 1,538 &757&781& 491 & 7,298&3,636&3,662\\ 
 LLMG & 656 & 3,936 &1,968&1,968& 165 & 990 &495&495& 822 & 4,932&2,466&2,466\\ 
 Total & 1,796 & 12,429 &5,646&6,783& 452 & 3,154 &1,440&1,714& 2,250 & 15,378&7,033&8,345\\ 
 \bottomrule
\end{tabular}
}
\caption{Data distributions between train, dev, and test splits.}
\label{table:splits}
\end{table*}

\begin{table*}[htb!]
\begin{center}
\scalebox{1}{%
\begin{tabular}{l|cccc|cccc|cccc}
\toprule
\multirow{2}{*}{} & \multicolumn{4}{c|}{BERT} & \multicolumn{4}{c|}{T5} & \multicolumn{4}{c}{LLaMa}\\  
& {CMV} & {Debate} & {LLMG} & {Avg} & {CMV} & {Debate} & {LLMG} & {Avg} & {CMV} & {Debate} & {LLMG} & {Avg}\\
\midrule
CMV &$0.789$ & $0.477$ & $0.557$ & $0.608$ & $0.771$ & $0.489$ & $0.752$ & $0.679$ & $0.820$ & $0.513$ & $0.652$ & $0.662$\\
Debate &$0.765$ & $0.594$ & $0.738$ & $0.699$ & $0.529$ & $0.616$ & $0.791$ & $0.645$ & $0.665$ & $0.672$ & $0.822$ & $0.720$\\
LLMG &$0.554$ & $0.473$ & $0.770$ & $0.599$ & $0.593$ & $0.483$ & $0.832$ & $0.636$ & $0.515$ & $0.453$ & $0.759$ & $0.576$\\
ALL &$0.760$ & $0.594$ & $0.727$ & $0.694$ & $0.782$& $0.632$& $0.933$& $0.782$& $0.851$ & $0.733$ & $0.834$ & $0.806$\\
\bottomrule
\end{tabular}%
}%
\end{center}
\caption{Cross-dataset finetuning performance of macro-F1 for three models. The row header represents the dataset that is used for training. ALL means training with all three datasets.}
\label{table:fine-tune}
\end{table*}

\section{Fully Supervised Finetuning}
\label{sec:supervised finetuning}

We now address \textbf{RQ1} and \textbf{RQ2} by conducting experiments with both traditional machine learning models and LLMs .

\subsection{Traditional Machine Learning Models}

We conduct experiments using widely adopted stance classification  methods. \citet{aldayel-2021-stance} identify SVM, CNN, and BiLSTM as leading machine learning methods for stance classification. Therefore, we conduct experiments to evaluate the effectiveness of LLMG as a weakly supervised approach for real-world datasets. 
We adopt Word2Vec embeddings---the well known embedding approach in the literature \citep{aldayel-2021-stance}. For BiLSTM and CNN, we fine-tune the models with a learning rate of $2\mathrm{e}{-4}$, AdamW optimizer, $0.5$ dropout, and CrossEntropy loss.

Table~\ref{table:baseline} presents the performance of classifying the three stance labels with and without LLMG as weak supervision. 
The results show that LLMG greatly improves the performance, particularly for CMV's Favor label and Debate's Against label, with average performance gains of 10\% and 9\%, respectively.
Our experiments demonstrate that incorporating AI-generated data enhances stance classification generalizability \citep{ng-2022-my}, reaffirming the benefits of LLMs in real-world applications \citep{lee-2022-coauthor, ravi-2024-small}.

Moreover, Table~\ref{tab:cohens_d} presents the Cohen's d effect sizes across the three models on two datasets with respect to three stance categories: Against, Favor, and None. 
The results suggest varying levels of statistical significance, where higher values correspond to stronger effects.
The SVM and CNN models are averaged over ten-fold cross-validation, while BiLSTM was trained for ten epochs.
Similarly, CNN and BiLSTM models show significant performance, especially in the Favor category for both datasets, with effect sizes ranging from 0.78*** to 1.31** across models.
Overall, these results indicate that incorporating LLMG (Language Model Guidance) is effective in improving stance classification, as evidenced by the strong effect sizes, particularly in distinguishing stances that favor a position.

\subsection{Large Language Models}

We term the previous generation of language models, such as BERT \citep{devlin-etal-2019-bert}, SLMs to contrast with current LLMs. 
A prevalent method for classification using SLMs involves \emph{finetuning}, which entails exposing a pretrained SLM to domain-specific data. 
However, finetuning is not always the optimal method for customizing LLMs and some research have suggested it could be detrimental to performance. Moreover, whereas finetuning is tractable for SLMs, it demands substantial computational resources for LLMs. 
Therefore, wepare SLMs and LLMs for supervised finetuning for stance classification.
For finetuning BERT, we concatenate the topic and argument with the special token [SEP] and prepend the sequence with the special token [CLS] to form the template [CLS] + Topic + [SEP] + Argument. A three way classification head is added on top of the token [CLS] to perform the classification task. 

For finetuning the generative models T5 and LLaMa, We use the same template as in the Training Prompt (below). We show a few concrete examples in Appendix~C. As for autoregressive pretraining, we apply the maximum likelihood estimation, which involves minimizing the cross-entropy loss between the predicted probability distribution of the next token and the actual token for the whole sequence. At inference time, we simply remove the gold label from the prompt so that the model can make a prediction. The output length is limited to 2. 

\begin{tcolorbox}[colback=black!5!white,colframe=gray!150!black,title=Training Prompt for Generative Models, coltitle = black!20!black]

Classify the stance of the argument towards the topic as either \emph{favor}, \emph{against}, or \emph{neutral}. Return the label only without any other text. 
\\

Topic: \{topic\}

Argument: \{argument\}

Label: \{label\}
\end{tcolorbox}

\subsection{Experimental Setup}

Our evaluation involves (1) BERT \citep{devlin-etal-2019-bert}, recognized for its effectiveness in classification, (2) T5 \citep{10.5555/3455716.3455856}, a generative counterpart to BERT, and (3)  We conduct the following experiments.

\paragraph{Finetuning with a single dataset} 
To evaluate generalizability in stance classification, we assess how a model trained on one dataset performs on another dataset.

\paragraph{Finetuning with multiple datasets} 
We extend the above evaluation to include finetuning on combined datasets. 

\paragraph{Finetuning with varied sizes of training data} 
We evaluate the effect of data size (from combined data) on finetuningo
For all datasets, we adopt the macro-F1 metric, namely, the average F1 score for each label category (Favor, Against, None). 
For BERT (110M) and T5 (250M), we perform finetuning with all parameters. 
For LLaMa-7b, we apply the QLoRA \citep{dettmers2023qlora} quantization technique, updating only 20 million parameters. 
We split the data into train, dev, and test sets. 
The data distribution is shown in Table~\ref{table:splits} below. 
The hyperparameter settings for all models are shown in Appendix~\ref{app:hyperparameter}.

\begin{figure*}[ht]
\centering
\includegraphics[scale=0.47]{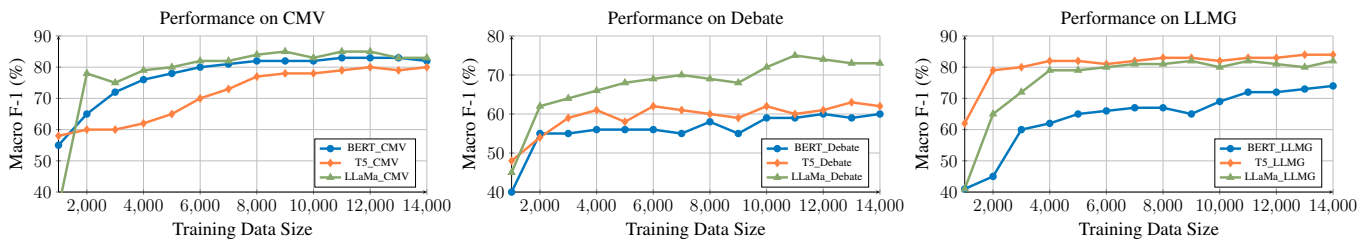}
\caption{Performance of micro-F1 for BERT, T5, LLaMa on all  partitions of the dataset with various training examples.}
\label{fig:fine-tune}
\end{figure*}

\subsection{Results}
We now present the results of our experimental investigations. 
Table~\ref{table:fine-tune} shows the results for finetuning with a single dataset and with all of the three datasets.  These results reveal a persistent challenge across all models: a difficulty in adapting to new datasets when subjected to finetuning with one dataset, indicating the subtle differences between domains.  Notably, the best average performance is achieved by LLaMa-7b fine-tuned on the Debate dataset.

For finetuning with multiple datasets, LLaMa-7b is the model with the highest average F1 across all three datasets. Despite having fewer fine-tuned parameters (20M compared to 110M and 250M), LLaMa-7b outperforms its counterparts, reflecting the power of LLMs in complex tasks. Both T5 and LLaMa-7 beat BERT, highlighting the advantage of using generative models over classification-oriented models for stance classification.

\paragraph{Ablation studies}
Figure~\ref{fig:fine-tune} presents the results for different amounts of training data. 
The three models demonstrate comparable and high training sample efficiency. Notably, with approximately 25\% of the training data, each model achieves nearly 95\% of its optimal performance.

\section{Zero-Shot and Few-Shot Benchmarking}
\label{sec:llm evaluation}

The zero-shot and cross-topic variants of stance classification are well-aligned since both rely on performing on topics not encountered during training. To address \textbf{RQ3}, we evaluate strict zero-shot and few-shot learning. Research suggests that the knowledge that LLMs possess is predominantly acquired through pretraining \citep{cruickshank-2023-use}. This implies that LLMs possess the inherent capacity to address various tasks, provided they are instructed in a suitable manner.

\subsection{Experimental Setup}
We focus on open LLMs as opposed to closed, commercially oriented LLMs \citep{touvron2023llama}, to enhance accessibility. LLMs exhibit a variety ofarchitectures and sizes, and whether they underwent instruction tuning during their training process. We employ LLaMa as the cornerstone of our study, because of its demonstrated superiority across multiple tasks and performance that is competitive with ChatGPT. We consider the 7B, 13B, 33B, 65B configurations of LLaMa, as well as the 7B, 13B, 33B configurations of its instruction-tuned counterpart, Vicuna \citep{vicuna2023}. We also include another model family, UL2 \citep{tay2023ul2}, and its instruction-tuned counterpart FLAN-UL2, which has an encoder-decoder architecture. We also include the 7B and 40B configurations of the Falcon family, with and without instruction tuning. 

We conduct experiments with zero-shot and few-shot in-context learning. For all the LLMs, we use QLoRA to quantize them to 4 bits to reduce the need for GPU memory. 
QLoRA suffers minimal loss on a variety of tasks \citep{dettmers2023qlora}. 
Our experiments are run on a mixture of NVIDIA-A100, NVIDIA-A30, NVIDIA-A10, and NVIDIA-A6000 GPUs.

\begin{table*}[htb!]
\begin{center}
\scalebox{0.84}{%
\begin{tabular}{l ccc ccc ccc ccc}\toprule
\multirow{2}{*}{\bf Models} & \multicolumn{3}{c }{\bf 0-shot} & \multicolumn{3}{c}{\bf 3-shot} & \multicolumn{3}{c}{\bf 6-shot} & \multicolumn{3}{c}{\bf 9-shot} \\ \cmidrule(lr){2-4} \cmidrule(lr){5-7}\cmidrule(lr){8-10}\cmidrule(lr){11-13}
& {\bf CMV} & {\bf Debate} & {\bf LLMG} & {\bf CMV} & {\bf Debate} & {\bf LLMG} & {\bf CMV} & {\bf Debate} & {\bf LLMG} & {\bf CMV} & {\bf Debate} & {\bf LLMG} \\\midrule
LLaMA-7B & 16.28& 16.33 & 15.08 & $17.26_{5.68}$ & $29.07_{6.10}$ & $24.45_{6.18}$ & $23.46_{8.15}$&$33.80_{6.33}$ & $29.32_{5.08}$ & 14.79 & 28.27& 31.66\\
LLaMA-13B & 9.68 & 20.48& 17.19 & $24.07_{7.56}$ &$35.49_{5.08}$& $31.84_{8.08}$ & $22.96_{6.99}$ & $41.34_{1.34}$ & $31.60_{3.79}$& 12.55& 40.84&30.50\\
LLaMA-30B & 22.05 & 11.58 & 38.23 & $32.63_{5.63}$& $42.68_{3.01}$ & $42.00_{2.02}$ & $28.68_{2.15}$ & $44.06_{1.68}$ & $37.03_{1.62}$ & 32.58& 42.10 & 39.97\\
LLaMA-65B & 33.32 & 36.96 & \textbf{54.59} & $38.40_{4.71}$ & $49.76_{2.38}$& $50.26_{3.31}$&$37.25_{5.69}$ &$50.21_{2.24}$ & $45.07_{2.55}$& 41.76& 50.73&53.65\\
Vicuna-7B & 24.38 & 37.49 & 33.05 &$19.83_{7.71}$ &$28.92_{5.61}$ &$29.94_{6.75}$& $15.95_{4.57}$& $30.63_{6.29}$& $30.70_{4.58}$& 12.15&26.10 &22.84 \\
Vicuna-13B & 25.59& 30.15& 47.11 & $31.29_{6.27}$& $41.31_{6.47}$& $38.90_{7.47}$& $22.90_{2.23}$&$42.41_{1.22}$& $36.78_{4.19}$& 14.95&42.56 &40.59 \\
Vicuna-33B & 34.51 & 42.59 & 51.86 & $31.70_{6.49}$& $47.54_{4.80}$& $41.40_{2.70}$& $29.92_{4.19}$&$48.94_{2.37}$& $38.50_{2.15}$& 29.48 & 49.90&41.50\\
Falcon-7B & 16.40 & 22.04 & 18.45 & $27.58_{3.60}$& $30.33_{7.68}$& $30.80_{1.84}$& $26.70_{3.88}$&$34.40_{5.45}$& $32.55_{3.80}$& 23.44&40.74 &34.59\\
Falcon-7B-I & 24.68 & 23.50 & 31.64 & $24.42_{2.03}$& $23.70_{3.44}$& $20.81_{1.73}$& $10.67_{1.38}$&$20.29_{0.57}$& $18.08_{1.29}$& 10.18& 20.63& 19.86\\
Falcon-40B & 30.52 & 39.20 & 33.85 & $24.73_{3.05}$& $34.54_{6.42}$& $28.87_{6.44}$& $22.01_{5.59}$&$32.67_{6.05}$& $29.22_{5.44}$& 16.34 &29.82 &23.61 \\
Falcon-40B-I & 35.22 & 42.52 & 38.74 & $29.87_{4.23}$& $35.07_{5.43}$& $32.74_{6.36}$& $27.60_{5.25}$&$33.07_{4.27}$& $31.44_{3.76}$& 22.75 &26.73 &27.74\\
UL2-20B & 31.93 & 37.32 & 36.38 & $27.49_{3.22}$& $22.35_{3.26}$& $29.36_{8.94}$& $25.54_{1.87}$&$18.28_{4.57}$& $19.91_{4.23}$& 22.59 &11.67 &17.01\\
FLAN-UL2-20B & \textbf{41.17} & \textbf{51.51} & 50.21 & \textbf{$41.85_{0.84}$}& \textbf{$52.51_{0.29}$}& \textbf{$52.70_{0.55}$}& $42.30_{0.63}$&$52.67_{0.31}$& $53.46_{0.62}$& 42.93 &52.60 &54.28\\
\bottomrule
\end{tabular}%
}%
\end{center}
\caption{Zero-shot and few-shot in-context learning for various LLMs.}
\label{table: llms}
\end{table*}

\subsection{Results}
The overall results are shown in Table~\ref{table: llms}. The main findings are summarized as follows. 

\paragraph{Significance test for model performance} Before starting our analysis, we performed McNemar’s test to assess the significance of model prediction differences among the top-performing models. This test was conducted in two stages. First, we performed an inter-model comparison, evaluating the top model from each family, namely LLaMA-65B, Vicuna-33B, Falcon-40B-instruct, and FLAN-UL2-20B. Second, we examined intra-model differences by comparing zero-shot and 9-shot performances within these four models. The corresponding p-values are presented in Table~\ref{table:pairwise significant test}. As observed, the differences across both settings are statistically significant (p < 0.001) across three datasets, with the exception of LLaMA-65B vs. Vicuna-33B on the CMV and LLMG datasets, and LLaMA-65B vs. FLAN-UL2-20B on the CMV dataset. Notably, all models exhibited significant differences between the zero-shot and 9-shot conditions, highlighting the critical role of few-shot examples in performance improvement.

\begin{table}[t]
\centering
\scalebox{0.8}{%
\begin{tabular}{l r r r}
\toprule
 & CMV & Debate & LLMG\\ 
 \midrule
LLaMa-65B vs. Vicuna-33B & 0.544 & <0.001 & 0.408 \\
LLaMa-65B vs. Falcon-40B-I & <0.001 & <0.001 & <0.001 \\
LLaMa-65B vs. FLAN-UL2-20B & 0.934 & <0.001 & <0.001\\
Vicuna-33B vs. Falcon-40B-I & <0.001 & <0.001 & <0.001\\
Vicuna-33B vs. FLAN-UL2-20B  & 0.671 & <0.001 & <0.001\\
Falcon-40B-I vs. FLAN-UL2-20B & <0.001 & <0.001 & <0.001\\
\midrule
LLaMa-65B 0-shot vs. 9-shot& <0.001 & <0.001 & <0.001\\
Vicuna-33B 0-shot vs. 9-shot& <0.001 & <0.001 & <0.001\\
Falcon-40B-I 0-shot vs. 9-shot & <0.001 & <0.001 & <0.001\\
FLAN-UL2-20B 0-shot vs. 9-shot & <0.001 & <0.001 & <0.001\\
 \bottomrule
\end{tabular}
}%
\caption{p-value for McNemar’s significant test. }
\label{table:pairwise significant test}
\end{table}

\begin{table}[t]
\centering
\scalebox{1}{%
\begin{tabular}{l r r r}
\toprule
 & CMV & Debate & LLMG\\ 
 \midrule
against argument & 39.54 & 60.24 & 35.01 \\
favor argument & 54.93 & 69.75 & 47.31\\
overall & 44.17 & 65.01 & 40.87 \\
 \bottomrule
\end{tabular}
}
\caption{Percentage of using facts in argument for each dataset. }
\label{table:facts}
\end{table}

\paragraph{Gap with upper bound} 
Overall, we observe positive effects for model scaling. For all model families, larger models yield better performances across most settings. 
However, the best performance model of FLAN-UL2, which achieves 41.17, 51.51, and 50.21 under zero shot for CMV, Debate, and LLMG, respectively, falls far behind the supervised approach, which suggests difficulty for LLMs to comprehend downstream tasks. 

\paragraph{Number of few-shot exemplars}
Exposing a model to more examples reliably improves performance across various tasks. However, our results are mixed. To understand variability, we randomly sampled 10 sets of examples for both 3-shot and 6-shot learning and calculated their mean and standard deviation. Some sets of examples show better performance than zero-shot. Nonetheless, the variability highlights the sensitivity of LLMs to specific examples. One exception is FLAN-UL2, the top performer, which maintains an average variance of 0.54, showcasing the consistency of its performance. Additionally, FLAN-UL2 demonstrates a robust improvement due to the increase in example input.

\paragraph{Impact of instruction tuning} 
Instruction-tuning is an approach to continually fine-tune an LLM by exposing it to diverse instructions and their responses. Doing so enhances the ability to follow user-specified instructions. Our results demonstrate that models fine-tuned through instruction reliably outperform models of the same architecture and size that are not instruction-tuned. This is apparent by comparing Vicuna to LLaMa, Falcon-instruct to Falcon, and FLAN-UL2 to UL2. This observation highlights the effectiveness of instruction-tuning as a task-agnostic method for adapting LLMs to downstream tasks.

\section{Conclusion}
\label{sec:conclusion}

In this paper, we compile our benchmark from three distinct sources: a social media platform, two debate websites, and arguments generated by large language models (LLMs). The resulting dataset covers a wide range of 4,498 distinct topics, comprising 30,961 arguments distributed across 21 diverse domains. 

We employ three distinctive experimental approaches: fully supervised learning, zero-shot learning, and few-shot in-context learning with LLMs to demonstrate the usefulness of our dataset. Notably, our findings highlight the superior performance of generative models over classification modelss, when used in a zero-shot scenario, demonstrate commendable performance, though with a noticeable performance gap relative to the upper bound. Furthermore, tuned LLMs reliably outperform their non-instruction-tuned counterparts, emphasizing the effectiveness of instruction-tuning for adapting LLMs to downstream tasks.

In conclusion, our study establishes robust baselines for the created dataset and provides valuable insights that can guide the development of more generalized stance classification methods. This research not only advances our understanding of the performance dynamics among different learning approaches but also offers practical implications for optimizing the use of LLMs in stance classification

\section{Discussion}
\label{sec:discussion}

We investigate the use of factual information in argument formulation in our benchmark and the limitation of our work. 

\subsection{Factual Information in Arguments}
The use of facts to build arguments improves credibility and makes them more persuasive by offering solid evidence to back up claims. Without facts, arguments often depend on personal opinions or emotions, which can lessen their effectiveness. The stance classification task primarily aims to infer the position or stance based on the content of an argument, without being directly concerned with the factual accuracy of the argument itself. However, in real-world applications, ensuring the factuality of arguments generated by LLMs is essential for their effective and responsible use.

We leverage an LLM distinct from the GPT family to facilitate the analysis of factual information prevalence in arguments. Specifically, we use \textit{Claude 3.5 Sonnet} from Anthropic with the facts extraction prompt. We choose this model as it is one of most advanced commercial models. 

\begin{tcolorbox}[colback=black!5!white,colframe=gray!150!black,title=Facts Extraction Prompt, coltitle = black!20!black]

Facts are objective statements that are verifiable. Arguments are subjective claims or positions.
\\

Given a topic and an argument, identify if the argument relies on any verifiable facts. 
Return the fact that the argument relies on. 
Return none if the argument does not rely on verifiable facts. Be concise in your response.
\\

Topic: \{topic\}

Argument: \{argument\}
\end{tcolorbox}

We calculate the percentage of using facts in arguments for CMV, Debate, and LLMG. The results are shown in Table~\ref{table:facts}. There is a clear difference between the use of facts in CMV discussions and formal debate arguments. CMV is a more casual platform for online users, while debate arguments come from structured debating websites. The latter stresses the importance of using factual evidence to back up claims, showing that including facts can \ improve the strength of an argument. In contrast, arguments produced by the GPT-3 model show a lower frequency of using factual information, which may be due to limitations set by built-in guardrails within the model. Across all three datasets examined, the frequency of factual information in \textit{favor} arguments is much higher than that in \textit{against}.

\subsection{Broader Impact}

The stance classification task focuses on identifying the viewpoint or position of an argument regarding specific topics, rather than evaluating the factual correctness of the information in the argument. While factual accuracy may not be the main concern of this task, it is important to note that text generated by LLMs or online users can sometimes include inaccurate or false information. This concern is particularly significant when our dataset is employed for purposes such as argument generation or augmentation, which is a potential application of our dataset.

Arguments containing unverified information in our dataset may lead to the creation of models that generate unreliable outputs, which can contribute to the spread of misinformation \citep{bang-2023-multitask}. The existing research on factual information validation can be categorized into two primary approaches: those that utilize external sources and those that do not. One approach focuses on extracting atomic facts from content generated by LLMs and validating these facts against external references, such as Wikipedia or Google search \cite{chern2023factoolfactualitydetectiongenerative} results. The alternative approach relies on the model's internal mechanisms, employing strategies such as chain-of-thought reasoning to deliberate on the generated responses and implement self-correction \cite{dhuliawala-etal-2024-chain}.

This study concentrates on stance classification and does not specifically examine the issue of factual accuracy in the collected or generated content. The risk of misinformation highlights the necessity for the establishment of rigorous evaluation metrics to assess the factual accuracy of AI-generated text \citep{shafayat-2024-multi} . 

\subsection{Limitations and Future Work}
This study faces several limitations. First, our proposed framework for the collection and orchestration of diverse argumentative sentence pairs, covering a wide array of topics, can be extended as needed to facilitate the collection of additional data. However, this framework is constrained by the types of platforms from which stance labels can be extracted. While we investigate the utilization of LLMs to construct stance classification datasets, more sophisticated experiments would be beneficial for exploring the full potential of this approach.

Second, while our study examines stance classification from three distinct sources, it is important to recognize that this task is applicable in a much boarder array of contexts, such as news articles, tweets, and political discourse. Therefore, combining our dataset with other existing datasets from different domains could improve the generalizability of stance classification. 

Third, while we focus on the adoption of GPT-3 for generating arguments, we do not directly compare LLMs for generating arguments. We defer comparative studies involving LLMs such as GPT-4, PaLM, or Claude to future research. Such an evaluation would enable more robust methods for benchmarking for stance classification and other social media problems.

\bibliography{Jiaqing, Ruijie}

\clearpage
\section*{Paper Checklist to be included in your paper}

\begin{enumerate}

\item For most authors\ldots 
\begin{enumerate}
    \item  Would answering this research question advance science without violating social contracts, such as violating privacy norms, perpetuating unfair profiling, exacerbating the socio-economic divide, or implying disrespect to societies or cultures?
    \answerYes{Yes}
  \item Do your main claims in the abstract and introduction accurately reflect the paper's contributions and scope?
    \answerYes{Yes}
   \item Do you clarify how the proposed methodological approach is appropriate for the claims made? 
    \answerYes{Yes}
   \item Do you clarify what are possible artifacts in the data used, given population-specific distributions?
    \answerNo{No, the dataset doesn't involve population-specific distributions}
  \item Did you describe the limitations of your work?
    \answerYes{Yes}
  \item Did you discuss any potential negative societal impacts of your work?
    \answerNo{No}
      \item Did you discuss any potential misuse of your work?
    \answerNo{No}
    \item Did you describe steps taken to prevent or mitigate potential negative outcomes of the research, such as data and model documentation, data anonymization, responsible release, access control, and the reproducibility of findings?
    \answerNo{No}
  \item Have you read the ethics review guidelines and ensured that your paper conforms to them?
    \answerYes{Yes}
\end{enumerate}

\item Additionally, if your study involves hypotheses testing\ldots 
\begin{enumerate}
  \item Did you clearly state the assumptions underlying all theoretical results?
    \answerNA{NA}
  \item Have you provided justifications for all theoretical results?
    \answerNA{NA}
  \item Did you discuss competing hypotheses or theories that might challenge or complement your theoretical results?
    \answerNA{NA}
  \item Have you considered alternative mechanisms or explanations that might account for the same outcomes observed in your study?
    \answerNA{NA}
  \item Did you address potential biases or limitations in your theoretical framework?
    \answerNA{NA}
  \item Have you related your theoretical results to the existing literature in social science?
    \answerNA{NA}
  \item Did you discuss the implications of your theoretical results for policy, practice, or further research in the social science domain?
   \answerNA{NA}
\end{enumerate}

\item Additionally, if you are including theoretical proofs\ldots 
\begin{enumerate}
  \item Did you state the full set of assumptions of all theoretical results?
    \answerNA{NA}
	\item Did you include complete proofs of all theoretical results?
    \answerNA{NA}
\end{enumerate}

\item Additionally, if you ran machine learning experiments\ldots 
\begin{enumerate}
  \item Did you include the code, data, and instructions needed to reproduce the main experimental results (either in the supplemental material or as a URL)?
    \answerNo{No, all material will be published later }
  \item Did you specify all the training details (e.g., data splits, hyperparameters, how they were chosen)?
    \answerYes{Yes}
     \item Did you report error bars (e.g., with respect to the random seed after running experiments multiple times)?
    \answerYes{Yes}
	\item Did you include the total amount of compute and the type of resources used (e.g., type of GPUs, internal cluster, or cloud provider)?
    \answerYes{Yes}
     \item Do you justify how the proposed evaluation is sufficient and appropriate to the claims made? 
    \answerYes{Yes}
     \item Do you discuss what is ``the cost'' of misclassification and fault (in)tolerance?
    \answerNo{No}
  
\end{enumerate}

\item Additionally, if you are using existing assets (e.g., code, data, models) or curating/releasing new assets, \textbf{without compromising anonymity}\ldots 
\begin{enumerate}
  \item If your work uses existing assets, did you cite the creators?
    \answerNo{No, no previous dataset is used.}
  \item Did you mention the license of the assets?
    \answerNo{No, no license is needed. The dataset will be open}
  \item Did you include any new assets in the supplemental material or as a URL?
    \answerNo{No}
  \item Did you discuss whether and how consent was obtained from people whose data you're using/curating?
    \answerNo{No}
  \item Did you discuss whether the data you are using/curating contains personally identifiable information or offensive content?
    \answerNo{No, no personally identity is involved}
\item If you are curating or releasing new datasets, did you discuss how you intend to make your datasets FAIR?
\answerNo{No}
\item If you are curating or releasing new datasets, did you create a Datasheet for the Dataset? 
\answerNo{No}
\end{enumerate}

\item Additionally, if you used crowdsourcing or conducted research with human subjects, \textbf{without compromising anonymity}\ldots 
\begin{enumerate}
  \item Did you include the full text of instructions given to participants and screenshots?
    \answerNA{NA}
  \item Did you describe any potential participant risks, with mentions of Institutional Review Board (IRB) approvals?
    \answerNA{NA}
  \item Did you include the estimated hourly wage paid to participants and the total amount spent on participant compensation?
    \answerNA{NA}
   \item Did you discuss how data is stored, shared, and deidentified?
   \answerNA{NA}
\end{enumerate}

\end{enumerate}

\appendix
\section{Websites for collecting topics}
\label{app:source}
\begin{enumerate}
    \item https://blog.kialo-edu.com/lesson-ideas/classroom-debate-ideas/
    \item https://www.weareteachers.com/controversial-debate-topics/
    \item https://research.com/education/debate-topics-for-college-students
    \item https://www.theedadvocate.org/political-debate-topics/
    \item https://owlcation.com/academia/100-Debate-Topics
    \item https://noisyclassroom.com/debate-topics/
    \item https://www.myspeechclass.com/controversial-speech-topics.html
    \item https://custom-writing.org/blog/debate-topics
    \item https://www.5staressays.com/blog/speech-and-debate/debate-topics
\end{enumerate}

\section{Experimental Hyperparameters}
We provide the main hyperparameter settings for data generation with GPT-3, and the training of BERT, T5, and LLaMa below. 
\label{app:hyperparameter}

\subsection{Argument Generation with GPT-3}
\begin{itemize}
    \item temperature: 0.7
    \item max\_generated\_tokens: 500
\end{itemize}

\subsection{BERT Finetuning}
\begin{itemize}
    \item Number of training epochs: 10
    \item Learning rate: 5e-5
    \item Batch size: 16
    \item Optimizer: Adamw
\end{itemize}

\subsection{T5 Finetuning}
\begin{itemize}
    \item Number of training epochs: 10
    \item Learning rate: 2e-5
    \item Batch size: 16
    \item Optimizer: Adamw
\end{itemize}

\subsection{LLaMa-7B Finetuning}
\begin{itemize}
    \item Number of training epochs: 3
    \item LoRA rank: 8
    \item LoRA alpha: 16
    \item LoRA dropout: 0.05
    \item LoRA bias: none
    \item Target modules: q\_proj, v\_proj, o\_proj, k\_proj, up\_proj, down\_proj, gate\_proj, embed\_tokens, lm\_head
    \item Learning rate: 2e-4
    \item Batch size: 64
    \item Warmup steps: 10
    \item Optimizer: paged\_adamw\_8bit
    \item max\_input\_length: 256
\end{itemize}

\subsection{LLaMa-7B Inference}
We perform inference with the fine-tuned LLaMa-7B in 4-bit with BitsAndBytes:
\begin{itemize}
    \item load\_in\_4bit: True
    \item bnb\_4bit\_quant\_type: nf4
    \item bnb\_4bit\_use\_double\_quantL: True
    \item bnb\_4bit\_compute\_type: bfloat16
    \item max\_generation\_length: 2
\end{itemize}

\section{Training Examples}
\label{app:training_examples}
We shown a few more randomly chosen examples from each datasets of the benchmark for training in Table~\ref{table:more_examples}. 

\begin{table*}[t]
\centering
\scalebox{0.9}{%
\begin{tabular}{p{3cm} p{12.5cm} l}
\toprule
Topical Claim & Comment & Stance\\ 
 \midrule
 Transgender (specifically mtf) athletes have an unfair advantage in strength-based sports and should only be able to participate in their biological gender group&This is in response to the post that hit the front page recently which can be seen here:   Biologically speaking, the male gender is already predisposed to stronger traits and as a response the Olympic Committee split competitions into two groups so that both genders have an \"equal playing field\". But with more acceptance and social tolerance, along with more transgender people coming out, problems have arose and will continue to rise until this gets dealt with and, to me, it's unfair and should be given strict guidelines and have no gender fluidity in the eyes of competitors. & Favor\\
 I feel that wealthy should pay a flat tax rather than the current progressive us tax system&I won't try to convince you that you're wrong about tax rates because I don't know enough about it.  Just consider: doesn't it ever bother you that people just pick up and leave for the sake of a little more money? These people have no ties their location, or they have ties that they're willing to throw away for a cheaper house. It's more of an admirable goal to build those deep, binding ties. Approaching the world with love, making friends, and investing in a community will lead to a more satisfying life than approaching it with cold logic centered around money. Just picture the end of your life for a moment. Will you be happier if you kept more of your money for yourself? Or will you be happier if you're a respected member of a community?   Find a place that's worth the higher taxes. Find a place you could fall in love with& Against\\
 It is not okay to keep animals in zoos.&Education and conservation efforts can be achieved through alternative means, such as wildlife sanctuaries and educational programs. These alternatives allow for a more ethical approach to animal welfare, as they focus on providing a natural environment for the animals while still educating the public about conservation and wildlife protection.& Favor\\
 The drinking age should be 18.&The human brain continues to develop until the mid-20s, particularly the prefrontal cortex responsible for decision-making and impulse control. Allowing 18-year-olds to consume alcohol may expose them to potential harm and hinder their brain development. Raising the drinking age to 21 provides additional years for the brain to mature and reduces the risk of long-term negative consequences associated with alcohol consumption.& Against\\
 Government surveillance is essential for national security.&In an increasingly interconnected world, a country's security can be threatened by various factors beyond direct self-defense. Issues such as terrorism, cyber attacks, and the spread of weapons of mass destruction pose significant risks that may require proactive military action. Waiting until an attack occurs could result in catastrophic consequences that could have been prevented.& None\\

 \bottomrule
\end{tabular}
}
\caption{More examples from our benchmark.}
\label{table:more_examples}
\end{table*}

\section{Complementary Results}
\label{app:per_class}
In this section, we provide complementary results for Table~\ref{table:fine-tune}. Specifically, per-class F1 score for each finetuning setting is shown in Table~\ref{table:BERT}, Table~\ref{table:T5}, and Table~\ref{table:LLaMa} for BERT, T5, and LLaMa, respectively. 

\begin{table*}[!htb]
\centering
\scalebox{0.75}{%
\begin{tabular}{lrrrrrr}
\toprule
\multirow{2}{*}{} & \multicolumn{2}{c}{\textbf{CMV}} & \multicolumn{2}{c}{\textbf{Debate}} & \multicolumn{2}{c}{\textbf{LLMG}}\\
\cmidrule(lr){2-3} \cmidrule(lr){4-5} \cmidrule(lr){6-7}
& {Against} & {Favor} & {Against} & {Favor} & {Against} & {Favor}\\
\midrule
CMV    & $0.738$ & $0.840$ & $0.559$ & $0.219$ & $0.622$ & $0.236$ \\
Debate & $0.658$ & $0.845$ & $0.733$ & $0.421$ & $0.660$ & $0.547$ \\
LLMG   & $0.272$ & $0.573$ & $0.357$ & $0.475$ & $0.631$ & $0.769$ \\
ALL   & $0.719$ & $0.747$ & $0.734$ & $0.443$ & $0.728$ & $0.701$ \\
\bottomrule
\end{tabular}
}%
\caption{Per class F1 score for BERT finetuning. }
\label{table:BERT}
\end{table*}

\begin{table*}[!htb]
\centering
\scalebox{0.75}{%
\begin{tabular}{lrrrrrr}
\toprule
\multirow{2}{*}{} & \multicolumn{2}{c}{\textbf{CMV}} & \multicolumn{2}{c}{\textbf{Debate}} & \multicolumn{2}{c}{\textbf{LLMG}}\\
\cmidrule(lr){2-3} \cmidrule(lr){4-5} \cmidrule(lr){6-7}
& {Against} & {Favor} & {Against} & {Favor} & {Against} & {Favor}\\
\midrule
CMV    & $0.664$ & $0.724$ & $0.558$ & $0.249$ & $0.687$ & $0.582$ \\
Debate & $0.440$ & $0.563$ & $0.623$ & $0.477$ & $0.840$ & $0.786$ \\
LLMG   & $0.451$ & $0.581$ & $0.427$ & $0.512$ & $0.820$ & $0.902$ \\
ALL   & $0.668$ & $0.722$ & $0.634$ & $0.522$ & $0.952$ & $0.900$ \\
\bottomrule
\end{tabular}
}%
\caption{Per class F1 score for T5 finetuning. }
\label{table:T5}
\end{table*}

\begin{table*}[!htb]
\centering
\scalebox{0.75}{%
\begin{tabular}{lrrrrrr}
\toprule
\multirow{2}{*}{} & \multicolumn{2}{c}{\textbf{CMV}} & \multicolumn{2}{c}{\textbf{Debate}} & \multicolumn{2}{c}{\textbf{LLMG}}\\
\cmidrule(lr){2-3} \cmidrule(lr){4-5} \cmidrule(lr){6-7}
& {Against} & {Favor} & {Against} & {Favor} & {Against} & {Favor}\\
\midrule
CMV    & $0.788$ & $0.880$ & $0.451$ & $0.429$ & $0.622$ & $0.664$ \\
Debate & $0.625$ & $0.561$ & $0.683$ & $0.551$ & $0.833$ & $0.801$ \\
LLMG   & $0.461$ & $0.553$ & $0.441$ & $0.452$ & $0.828$ & $0.792$ \\
ALL   & $0.882$ & $0.852$ & $0.724$ & $0.643$ & $0.853$ & $0.821$ \\
\bottomrule
\end{tabular}
}%
\caption{Per class F1 score for LLaMa finetuning. }
\label{table:LLaMa}
\end{table*}

\end{document}